%% file: main.tex
\documentclass{bmvc2k}

\usepackage{graphicx}
\usepackage{booktabs}
\usepackage{multirow}
\usepackage{multicol}
\usepackage{comment}
\usepackage{tikz,pgfplots}
\usepackage{amssymb}
\newcommand{\squeezeup}{\vspace{-2.5mm}}

\title{\textcolor{teal}{FILS:} Self-Supervised Video \textcolor{teal}{F}eature Prediction \textcolor{teal}{I}n Semantic \textcolor{teal}{L}anguage \textcolor{teal}{S}pace}

\addauthor{Mona Ahmadian}{m.ahmadian@surrey.ac.uk}{1}
\addauthor{Frank Guerin}{f.guerin@surrey.ac.uk}{1}
\addauthor{Andrew Gilbert}{a.gilbert@surrey.ac.uk}{1}

\addinstitution{
 University of Surrey\\
 Guildford, UK
}

\runninghead{FILS}{Self-Supervised Video Feature
Prediction In Semantic Language Space}

\begin{document}

\maketitle

\begin{abstract}
\noindent This paper demonstrates a self-supervised approach for learning semantic video representations. Recent vision studies show that a masking strategy for vision and natural language supervision has contributed to developing transferable visual pretraining. Our goal is to achieve a more semantic video representation by leveraging the text related to the video content during the pretraining in a fully self-supervised manner. To this end, we present \textbf{FILS}, a novel self-supervised video \textbf{F}eature prediction \textbf{I}n semantic \textbf{L}anguage \textbf{S}pace (FILS). The vision model can capture valuable structured information by correctly predicting masked feature semantics in language space. It is learned using a patch-wise video-text contrastive strategy, in which the text representations act as prototypes for transforming vision features into a language space, which are then used as targets for semantically meaningful feature prediction using our masked encoder-decoder structure.
FILS demonstrates remarkable transferability on downstream action recognition tasks, achieving state-of-the-art on challenging egocentric datasets, like Epic-Kitchens, Something-SomethingV2, Charades-Ego, and EGTEA, using ViT-Base. Our efficient method requires less computation and smaller batches compared to previous works.
\end{abstract}


\section{Introduction}
\label{sec:intro}
\input{tex/intro.tex}

\section{Related Works}
\label{sec:related}
\input{tex/related.tex}

\section{Self-Supervised Video Feature Prediction in Semantic Language Space}
\label{sec:method}
\input{tex/method}

\section{Experiments}
\label{sec:exp} 
\input{tex/results}

\section{Conclusion}
\label{sec:conclusion} 
\input{tex/conclusion}

\bibliography{egbib}

\hfill \clearpage

\input{tex/supp}

\end{document}

%% file: tex/intro.tex
Self-supervised pretraining, mainly through masked reconstruction, has demonstrated significant success in natural language processing~\cite{devlin2019naacl,brown2020language} and more recently in video~\cite{tong2022videomae}. Additionally, using web text for self-supervision in visual learning has been promising~\cite{radford2021learning}, with applications extending to video~\cite{zhao2023learning}. Video, however, has some unique characteristics: dense data with frame-to-frame redundancy and crucial action segments pivotal for comprehension. Therefore, applying techniques borrowed from the image domain may be suboptimal due to a lack of focus on the essential part of the frame. Our challenge lies in effectively integrating recently developed self-supervision concepts such as masking and text supervision tailored for video representation.

Addressing text supervision, early approaches relied on labeled text from supervised datasets, restricting models to predefined categories and labels~\cite{wang2021actionclip}. Recent advancements have shifted towards leveraging Large Language Models (LLMs) and vision-language models for text supervision, either through a text bag~\cite{lin2023match} or dense captions~\cite{zhao2023learning}. To effectively apply text supervision, contrastive learning on related images and captions, such as CLIP, has proven effective in building powerful representations~\cite{radford2021learning}. 
Nevertheless, their counterparts in the video domain do not show the same generality~\cite{qian2022multimodal,wang2021actionclip} and do not consider temporality, a key aspect for understanding a video.

Defining the objective for reconstructing masked video poses a challenge. Tong et al.~\cite{tong2022videomae} initially performed reconstruction in pixel space with masked autoencoders. However, Tan et al.~\cite{tan2021vimpac} suggest reconstructing in the latent space of a quantized-variational autoencoder to avoid overfitting on low-level visual information. Similarly, Assran et al.~\cite{assran2023self} found that learning in a higher semantic space leads a model to learn more semantic features in images. 
Yang et al.~\cite{yang2023rils} take this to language semantics, reconstructing image patches by mapping them to a distribution over textual features to predict the semantics of masked patches within a linguistic context. They employ image-text contrastive learning and incorporate a reconstructive loss. However, their random selection of image patches may not be optimal for video representation learning, where only certain parts of the scene capture the activity that should align with the video's text caption. If we concentrate contrastive learning on crucial video parts for activity comprehension, video-text alignment can happen more efficiently, emphasizing essential semantic features necessary for comprehensive video understanding.

The impressive progress in these research directions encouraged us to consider the following: Can we merge a masking strategy and language guidance to enhance visual pretraining? Self-supervised objectives can operate within a latent space established with language, maintaining the alignment of language with learned visual representations. This improves the interpretability and semantic aspects of the representations. A straightforward approach to achieving this objective is integrating a masking strategy with video-text contrastive learning for multi-task learning. Still, this simple combination cannot fully cover the potential synergies between these two objectives.

\squeezeup
\begin{figure}[htb]
  \centering
  \includegraphics[trim={0cm 0cm 0cm 0cm}, height=3 cm]{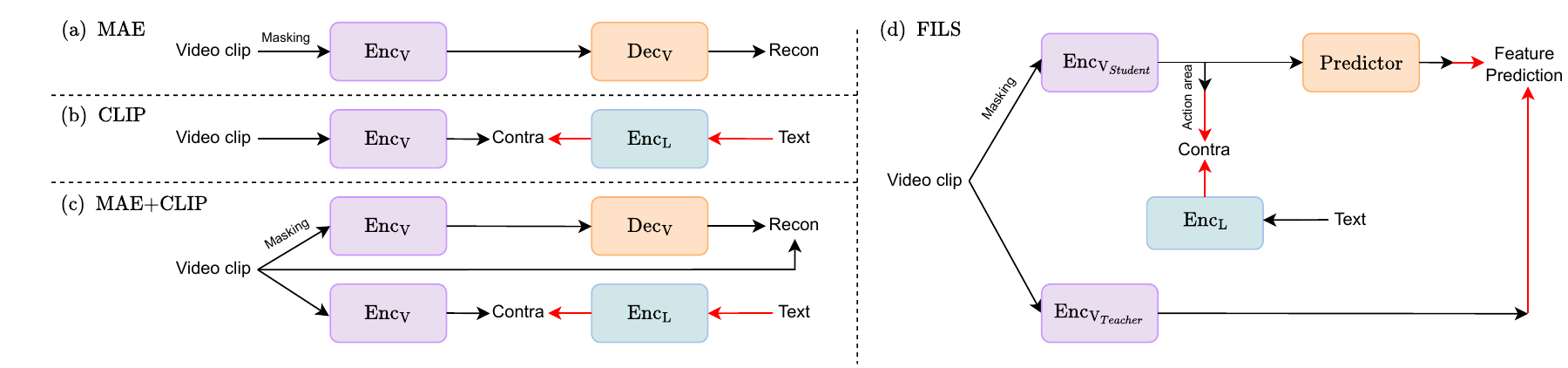}
  \vspace{3pt}
  \caption{Architecture comparisons between MAE,
CLIP, MAE+CLIP, and FILS. Contra indicates video-text contrastive loss. The red arrow points to the language space, while the black ones indicate the knowledge flow in the vision space.
  }
  \label{fig:archis}
\end{figure}

We address these combined issues by drawing on the best practices from related works and developing \textbf{FILS} to inherit the benefits of these approaches. FILS conducts video feature prediction in the semantic language space, completely self-supervised. We use natural descriptions provided by an off-the-shelf video caption model for text supervision~\cite{Yu_VideoBLIP} and train our video encoder in the CLIP language space~\cite{radford2021learning} employing a contrastive learning objective to learn high-level semantic features. We aim to predict masked video features within this shared space. We narrow our focus on the video regions where significant action unfolds by identifying \textit{action area} using a method that detects motion from optical flow but ignores camera motion~\cite{ahmadian2023mofo}. We introduce ActCLIP, which can exclusively utilize patches from these regions during contrastive learning, aligning them closely with language embeddings of the video captions. This strategy ensures that our model captures the essence of the depicted activities, emphasizing objects and actions relevant to the identified action areas. While contrastive learning benefits from larger batch sizes, our ActCLIP shows promise in surpassing the CLIP performance for video representation learning while maintaining batch sizes and computation requirements low. In Fig.~\ref{fig:archis}, you can see the semantic comparisons between MAE, CLIP, a simple masking strategy with video-text contrastive learning termed MAE+CLIP, and our proposed FILS. Our method outperforms all previous self-supervised learning techniques involving ViT-b and has excellent generalization capability for downstream tasks like action recognition. 
The following briefly describes our primary contributions:
\begin{itemize}
    \item A new self-supervised method for video representation learning from unlabeled videos and their captions; a feature prediction strategy on masked video and patch-wise contrastive learning within potential action areas that enriches our video encoder with richer, abstract representations. 
    \item In ActCLIP, we integrate motion's importance for action identification in videos by contrastive learning between patches in recognized action regions and associated textual context.
    \squeezeup
    \item  Demonstrating the effectiveness of using mutual information to prioritize patches that convey semantic and action-related details, enhancing the utility of learned representations across various tasks, including achieving better performance in action recognition through feature prediction-based visual representations.
\end{itemize}
\squeezeup
\squeezeup

%% file: tex/related.tex
\textbf{Video self-supervised learning}
Video understanding is a fundamental aspect of visual recognition, involving the analysis of video inputs. Video understanding tasks like action classification, detection, and object segmentation~\cite{tong2022videomae,ahmadian2023mofo,li2021ego,sun2019videobert,he2023clip} require handling complex spatial and temporal data. Rather than building representations using costly and limited human-defined annotations, self-supervised learning extracts discriminative video features from unlabeled data and avoids expensive annotations. The standard vision-language pretraining pipeline, which involves initial pretraining and finetuning afterwards, is designed to develop a general multimodal feature representation suitable for multiple downstream tasks~\cite{wei2022masked,tong2022videomae,baevski2022data2vec,assran2023self}. Recent generative self-supervised methods~\cite{he2022masked,wei2022masked,xie2022simmim} have advanced the field. Masked Auto-Encoders (MAEs)~\cite{he2022masked,girdhar2023omnimae,tong2022videomae} randomly mask input pixel patches and reconstruct them by minimizing reconstruction errors in pixel space, showing competitive performance when finetuned. Other work reconstructs directly in the latent space~\cite{assran2023self,baevski2022data2vec} or aims to predict contextualized latent representations containing information from the entire input through masked prediction~\cite{baevski2022data2vec}.

\noindent \textbf{Vision-language representation learning} 
In multimodal video-language learning and also language-guided video comprehension, incorporating language alongside videos~\cite{lei2021less,fu2021violet,zhang2021vinvl,sun2019videobert,tan2021vimpac,zhao2023learning,ashutosh2023hiervl} has introduced many intriguing challenges. Numerous attempts have been undertaken to merge computer vision and language, using the combined knowledge for various multimodal applications. Early work explored loss functions and architectures to grasp semantic vision-language alignments~\cite{wang2018learning,zheng2020dual}. Even before deep learning became popular, early research investigated the process of learning visual representations from image captions~\cite{quattoni2007learning}. Vision-language pretraining~\cite{lu2019vilbert,sun2019videobert,sun2019learning,radford2021learning} has become prevalent for learning transferable multimodal representations to enhance video-text tasks like captioning~\cite{yu2023efficient}, visual question answering~\cite{piergiovanni2022video}, and referring segmentation~\cite{yang2022lavt,yang2021hierarchical} and others. Before masked autoencoders, contrastive learning jointly learned vision and language representations. A notable example is Contrastive Language-Image Pretraining (CLIP)~\cite{radford2021learning}, which aligns image and text embeddings via contrastive loss, achieving supervised-comparable results. It used modality-specific encoders projecting to a shared embedding space with image-text pairs as targets. Work like~\cite{he2023clip} combined CLIP's language guidance with self-supervised contrastive learning for semantically-aligned pixel embeddings, boosting finetuning, and showcasing language's generality in supervising vision. Recent works~\cite{dong2023maskclip,yang2023rils,li2023scaling} explore incorporating masking into such pretraining, with strong results when finetuned on extensive labeled data.

%% file: tex/method.tex
\begin{figure}[tb!]
  \centering
  \includegraphics[trim={3cm 15.6cm 15cm 0cm}, height=4.7cm]{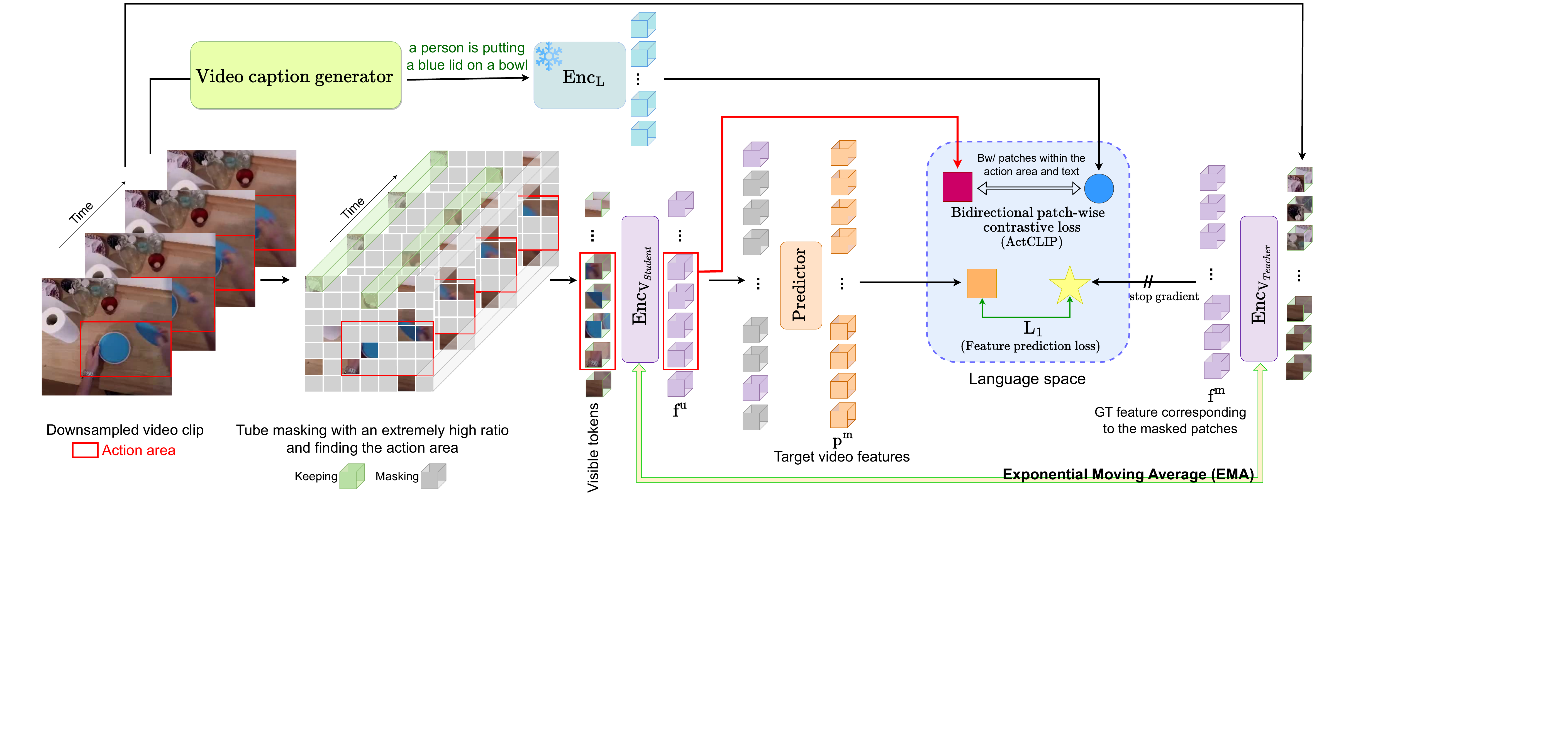}
  \caption{Overview of our method. We perform self-supervised feature prediction and video-text contrastive learning simultaneously. 
  The red arrow denotes the features of the patches within the action area.}
  \label{fig:FILS}
\end{figure}

We introduce FILS, a framework for deeper video understanding that leverages a unified embedding space that integrates video and natural language for multimodal learning. Fig.~\ref{fig:FILS} presents an overview of our method; it has two objectives: 1) Feature Prediction, where the input is masked and encoded (the student mode with random initialisation), and the predictor predicts representations. Then, we create representations of all the input data, which are meant to be targeted for the learning task (the teacher mode). As we will discuss in Sec.~\ref{teacher}, to prevent the collapse, the teacher tracks student parameters, and their weights are derived from the exponentially moving average of the student~\cite{he2020momentum,grill2020bootstrap}. 2) ActCLIP, an auxiliary CLIP-based self-supervised objective performing contrastive learning between motion or action area patches and relevant text, aligning video and language spaces to learn semantic context.
More information is in Sec.~\ref{subsec:training-obj}.

\subsection{Model Architecture}

\noindent\textbf{Video Encoder.}
In our approach, we cast video representation learning as a feature prediction objective involving comparing predicted features from the predictor and their corresponding ground truth features. To ensure stable representation during this task, we adopt a teacher/student approach. As is typical with self-supervised works~\cite{tong2022videomae,wang2023videomae}, we utilise an encoder ($\text{Enc}_{\text{V}_\mathit{Student}}$) to compute the representation of the masked input using the unmasked (remaining) patches. As tube modelling has demonstrated superior capability in capturing temporal and spatial information compared to frame modeling~\cite{tong2022videomae}, we utilise tube embedding for a video clip to further concentrate on spatiotemporal saliency. The input to the student encoder is masked with a high proportion of video ($V$) patches using spatiotemporal tube masking. After masking, the student vision encoder encodes the unmasked patches  ($V_u$), resulting in a $f^u$ embedding. The ground truth of masked patches in the video input ($V_m$) is encoded by a teacher encoder, resulting in a $f^m$ embedding. Eq.~\ref{eq:fearure_embed} formulates  the process of vision encoders:

\begin{equation}
\begin{gathered}
    f^u =  \text{Enc}_{\text{V}_\mathit{Student}}(V_u),\qquad u\in[1,N_u]\\
    f^m =  \text{Enc}_{\text{V}_\mathit{Teacher}}(V_m),\qquad m\in[1,N_m]
    \label{eq:fearure_embed}
\end{gathered}
\end{equation}

\noindent$m$ and $u$ demonstrate the patch index related to masked and unmasked patches. $N_u$ and $N_m$ indicate the number of unmasked (visible) patches and the number of masked patches in that order.

\noindent\textbf{Predictor.} This module is made up of transformer blocks. Using a learnable [MASK] token and the unmasked encoded feature $f_u$ of the masked video as inputs, it decodes or predicts the features of the masked patches from a masked view. This results in $p^m$, the predicted features of the masked input, where $N_u$ indicates the number of unmasked (visible) patches. 
\begin{equation}
p^m=\text{Predictor}({f}^u,\text{[MASK]}),\qquad u\in[1,N_u], 
\end{equation}

\noindent\textbf{Teacher Parameterization.}\label{teacher}
To avoid representation collapse, an exponentially moving average (EMA) of the student model parameters $\theta$, with the model's weights in target weight $\Delta$, is used to parameterize the teacher model to encode the video clip patches: $\Delta\leftarrow\tau\Delta + (1-\tau)\theta$. We employ a schedule for $\tau$ that increases this parameter linearly throughout the first $\tau_n$ updates, from $\tau_0$ to the target value $\tau_e$. The value is then maintained constant for the rest of the training. When the model is random at the beginning of training, this strategy leads to more frequent updates to the teacher; however, when suitable parameters have already been learned, the frequency of updates to the teacher decreases.

\noindent\textbf{Text Encoder.} 
To enable the integration of textual information with our vision encoder, we utilise a text encoder that converts textual inputs into a latent representation for joint processing. A stack of transformer layers tokenizes the representation of input text ($T$) into the global representation of the input texts, $h=\text{Enc}_\text{L}(T)$.

\subsection{Training Objectives} \label{subsec:training-obj}
We propose two loss functions in our frameworks: a contrastive video-to-text loss and a feature prediction loss. These losses are crucial in aligning video and text modalities within the language embedding space, fostering effective cross-modal understanding.

\noindent\textbf{ActCLIP.}\label{ActCLIP}
Recognising the importance of motion in self-supervised video techniques for action recognition, we integrate it through contrastive learning between patches within motion or action area and relevant text, resulting in a unified language-vision space fostering coherent content understanding. Motion areas are identified following an approach in~\cite{ahmadian2023mofo} to represent motion precisely and are termed action areas in our work.
Then, to learn the parameters of the shared vison-language embedding space, we leverage a patch-wise video-text contrastive loss between patches within the action area to align both modalities in a shared embedding space. Given this contrastive objective happens in the detected action area, we called this \textit{ActCLIP}. Initially, we take the mean-pooled video feature of patches within the detected motion area:

\begin{equation}
    \Bar{f}=\frac{1}{N}\displaystyle\sum_{i=1}^{N_a} f^a
\end{equation}
$a$ indicates patches inside the action area, $N_a$ is the number of patches in the detected action area, and $f^a$ is the encoded feature representation of the patches in $a$. The video features are further mapped by projection head to the language space and normalisation on them, and the text feature comes after:

\begin{equation}
\begin{gathered}
  z^V = \lVert\theta(\Bar{f})\rVert,\\
  z^T = \lVert h \rVert
  \label{eq:mapping}
\end{gathered}
\end{equation}
\noindent$\lVert.\rVert$ and $\theta(.)$ denote the normalisation operation and the projection head (mapping) from video to text, respectively.

The bidirectional contrastive loss between video and text can be represented as follows:
\begin{equation}
\begin{gathered}
    L_{V2T} = -\frac{1}{B} \displaystyle\sum_{i=1}^{B} \log\frac{exp(\langle z_i^V,z_i^T\rangle/ \sigma)}{ \sum_{j=1}^{B}exp(\langle z_j^V,z_j^T\rangle/ \sigma)},\\
     L_{T2V} = -\frac{1}{B} \displaystyle\sum_{i=1}^{B} \log\frac{exp(\langle z_i^T,z_i^V \rangle/ \sigma)}{ \sum_{j=1}^{B}exp(\langle z_j^T,z_j^V)\rangle/ \sigma)}  
\end{gathered}
\end{equation}

\noindent$\sigma$ is a learnable parameter that is cooperatively trained throughout the pretraining; B denotes the batch size, and i and j represent the index inside a mini-batch. The following is a formulation of the total loss of video-text contrastive learning:
\begin{equation}
  L_\mathit{ActCLIP} = \frac{1}{2}(L_{V2T}+L_{T2V})
  \label{eq:ActCLIP}
\end{equation}

\noindent\textbf{Feature Prediction in Language Space.}\label{subsec:FP} 
Our reconstruction space is built by taking a vision-language perspective. We perform masked visual reconstruction in this language space, using the text features as natural semantic information for the video patches. Therefore, given the encoded patch features from the Teacher encoder corresponding to the masked patches ($f^m$) and predicted patch features ($p^m$) (reconstructed ones), where $m$ is the index of the masked patch, we first map and normalise both features to the language feature space:

\begin{equation}
  \tilde{p}^m_i = \lVert\theta(p^m_i)\rVert, \quad \quad \tilde{g}^m_i = \lVert\theta(f^m_i)\rVert
  \label{eq:NormPatches}
\end{equation}
where $i$ indicates data in the batch and $\theta(.)$ represents the same vision mapping in Eq.~\ref{eq:mapping}. So, both the masked prediction and its corresponding target are mapped into this language semantic space. Our proposed loss function for the prediction is \textbf{F}eature \textbf{P}rediction (FP) loss, which is the average L1 distance between the mapped predicted patch-level representations via predictor $\tilde{p}^m$, and the mapped target (ground truth) patch-level representation coming from the Teacher encoder $\tilde{g}^m$; 
\squeezeup
\squeezeup
\begin{equation}
   L_\mathit{FP} = D(\tilde{p},\tilde{g}) =  \frac{1}{N_m}\displaystyle\sum_{i=1}^{N_m}\lVert(\tilde{p}_i,\tilde{g}_i)\rVert_1
  \label{eq:FPLoss}
\end{equation}

\noindent\textbf{Overall Objective Function.}
The final objective of FILS is a combination of video-text contrastive loss and feature prediction loss.
$\lambda_1$ and $\lambda_2$ adjust the weighting between our proposed contrastive loss(ActCLIP) and feature prediction loss(FP).
\begin{equation}
  L_\mathit{FILS} = \lambda_1 L_\mathit{ActCLIP} + \lambda_2 L_\mathit{FP}
  \label{eq:MergeLoss}
\end{equation}
\squeezeup

%% file: tex/results.tex

We assess the quality of the representations learnt by FILS on the challenging action recognition task, which is difficult to automate despite being easy for humans. Our experiments show FILS achieves superior performance across all metrics on the utilized action recognition dataset, demonstrating how vision self-supervision enhances the vision-language contrastive approach. Further analysis, dataset details, metrics, and implementation details are provided in the supplementary material.

\begin{table}[t!]
  \caption{ \textbf{Performance of action recognition on EK100 and SSV2.} FILS outperforms all prior works regarding action-level top-1 accuracy. In the table below, p-data and L mention pretraining data utilized the incorporation of language during the training, respectively.}
\vspace{0.2cm}

\label{tab:FT}
    \resizebox{5cm}{!}{
    \begin{minipage}[h]{0.5\linewidth}
    \raggedleft
  \centering
  \phantom{hellohello}(a) Epic-Kitchens
  \scalebox{0.76}{
  \begin{tabular}{c c c c ccc}
    \toprule
    \multirow{2}{*}{Method} & \multirow{2}{*}{Backbone} & \multirow{2}{*}{p-data} & \multirow{2}{*}{L} & \multicolumn{1}{c}{Verb} & \multicolumn{1}{c}{Noun} & \multicolumn{1}{c}{Action} \\
    
      & & & & Top-1 & Top-1 & Top-1 \\
    \midrule
    SlowFast~\cite{feichtenhofer2019slowfast} & ResNet101 & K400 & $\times$  & 65.6 & 50.0 & 38.5\\
    TSM~\cite{lin2019tsm} & ResNet50 & IN-1K & $\times$  & 67.9 & 49.0 & 38.3\\
    Mformer~\cite{patrick2021keeping} & ViT-L & IN-21K+K400 &$\times$ & 67.1 & 57.6 & 44.1\\
    Video Swin~\cite{liu2022video} & Swin-B & K400 & $\times$ & 67.8 & 57.0 & 46.1\\
    ViViT FE~\cite{arnab2021vivit}& ViT-L & IN-21k+K400 & $\times$& 66.4 &56.8& 44.0\\
    IPL~\cite{wang2021interactive} & I3D & K400 &$\checkmark$ & 68.6 & 51.2 & 41.0\\
    Omnivore~\cite{girdhar2022omnivore}  & Swin-B & IN+K400+SUN & $\times$& 69.5& 61.7& 49.9\\
    MeMViT~\cite{wu2022memvit} & ViT-B & K600 &$\times$ & 71.4& 60.3& 48.4\\
    MTV~\cite{yan2022multiview} & MTV-B & WTS-60M &$\times$ &69.9 &63.9& 50.5\\
    LaViLa~\cite{zhao2023learning} & TSF-B & WIT+Ego4D& $\checkmark$& 69.0 & 58.4 & 46.9\\
    AVION~\cite{zhao2023training} & ViT-B  & WIT+Ego4D& $\checkmark$& 70.0 & 59.8 & 49.1\\
     VideoMAE*~\cite{tong2022videomae} & ViT-B & EK100& $\checkmark$ & - & - &  48.5 \\   
    \textbf{FILS(ours)} & ViT-B & EK100& $\checkmark$ & 72.2 & 61.7& 51.0\\

  \bottomrule
  \end{tabular}
  }
      \label{tab:EK100}
    \end{minipage}  }
    \hspace{2cm}
    \resizebox{5cm}{!}{
    \begin{minipage}[h]{0.5\linewidth}
    \raggedleft
    \centering
    (b) Something\text{-}Something V2
    \scalebox{0.76}{
  \begin{tabular}{c c c c c c}
    \toprule
    Method & Backbone & p-data & L & Top-1\\
    \midrule
    SlowFast~\cite{feichtenhofer2019slowfast} & ResNet101 & K400  & $\times$  & 63.1\\
    TSM~\cite{lin2019tsm} & ResNet50 & K400 & $\times$  & 63.4\\
    TimeSformer~\cite{bertasius2021space} & ViT-L & IN-21K & $\times$ & 62.4\\
    Mformer~\cite{patrick2021keeping} & ViT-L & IN-21K+K400&$\times$ & 68.1\\
    Video Swin~\cite{liu2022video}& Swin-B & K400  & $\times$&69.6\\
    ViViT FE~\cite{arnab2021vivit}& ViT-L & IN-21k+K400 & $\times$& 65.9\\
    VIMPAC~\cite{tan2021vimpac} & ViT-L & HowTo100M & $ \checkmark $& 68.1\\
    BEVT~\cite{wang2022bevt} & Swin-B & IN-1K + K400 &$\times$  & 70.6 \\
    VideoMAE~\cite{tong2022videomae} & ViT-B & SSV2 & $\times$ & 70.8\\
    OmniMAE~\cite{girdhar2023omnimae} & ViT-B & IN-1K + SSv2 & $\times$ & 69.5\\
    Omnivore~\cite{girdhar2022omnivore} & Swin-B & IN-21k+K400 & $\times$& 71.4\\
    
    VideoMAE V2~\cite{wang2023videomae}& ViT-B& UnlabeledHybrid & $\times$& 71.2\\
    \textbf{FILS(ours)} & ViT-B & SSV2 & $ \checkmark $ & 72.1\\
  \bottomrule
  \end{tabular}
  }
      \label{tab:SSV2}
    \end{minipage}
    }

\end{table}

\begin{table}[t!]
  \caption{\textbf{Charades-Ego and EGTEA Action Recognition.} FILS achieves substantial improvements in this task, outperforming prior works, while Charades-Ego and EGTEA videos visually differ from EK100 videos, which FILS is pretrained on. 
The table shows p-data and L, referring to the pretraining data utilized and language incorporation during training. Note that the * model has been trained by us using the code provided by its authors.}
\vspace{0.2cm}

\label{tab:othersFT}
    \resizebox{4.8cm}{!}{
    \begin{minipage}[h]{0.42\linewidth}
    \raggedleft
  \centering
  \phantom{hellohell}(a) Charades-Ego
  \scalebox{0.7}{
  \begin{tabular}{c c c c c}
    \toprule
    Method & Backbone & p-data & L & mAP\\
    \midrule
    ActorObserverNet~\cite{sigurdsson2018actor} & ResNet-152 & Charades & $\times $ &20\\
    SSDA~\cite{choi2020unsupervised} & I3D & Charades-Ego& $\times$ &25.8\\
    Ego-Exo~\cite{li2021ego}  & SlowFast-R101  & Kinetics-400 & $\times$ &30.1\\
    EgoVLP~\cite{lin2022egocentric} & TSF-B & Ego4D & $\checkmark$ &32.1 \\
    HierVL-Avg~\cite{ashutosh2023hiervl} &ViT-Base& Ego4D&$ \checkmark $ & 32.6 \\
    HierVL-SA~\cite{ashutosh2023hiervl} & ViT-Base & Ego4D & $ \checkmark $& 33.8\\
    EgoVLPv2~\cite{pramanick2023egovlpv2} & TSF-B & EgoClip & $\checkmark$ & 34.1\\
    LaViLA~\cite{zhao2023learning} & TSF-B &  WIT+Ego4D & $ \checkmark $ &33.7\\

    \textbf{FILS(ours)} & ViT-Base & EK100 & $ \checkmark $ & \textbf{34.4} \\
  \bottomrule
  \end{tabular}
  }
    \label{tab:chrades}
    \end{minipage}  }
    \hspace{1.5cm}
    \resizebox{6.3cm}{!}{
    \begin{minipage}[h]{0.6\linewidth}
    \raggedleft
    \centering
    (b) EGTEA
    \scalebox{0.7}{
  \begin{tabular}{c c c c c c}
    \toprule
    Method & Backbone & p-data & L & Top-1 Acc. & Mean Acc.\\
    \midrule
    Li et al.~\cite{li2018eye} & I3D  & K400 & $\times$ & - & 53.30\\
    LSTA~\cite{sudhakaran2019lsta}  & ConvLSTM & IN-1k &  $\times$ & 61.86 & 53.00\\
    IPL~\cite{wang2021interactive} & I3D & K400 & $\checkmark$ & - & 60.15\\
    MTCN~\cite{kazakos2021little} & SlowFast  & K400+VGG-S  & $ \checkmark $& 73.59 & 65.87\\
    LaViLA~\cite{zhao2023learning}  & TSF-B & WIT+Ego4D & $ \checkmark $ & 77.45 & 70.12\\
    \textbf{FILS(ours)} & ViT-Base & EK100 & $ \checkmark $   & \textbf{78.48} & \textbf{71.20} \\

  \bottomrule
  \end{tabular}
  }
      \label{tab:EGTEA}
    \end{minipage}
    }
\vspace{-0.2cm}
\end{table}

\subsection{Action Recognition Task}

We assess the learned video representation by finetuning the video encoder for action classification. In line with previous studies~\cite{tong2022videomae,zhao2023learning}, after finetuning the video encoder, top-1 accuracy is reported on verbs, nouns, and actions for Epic-Kitchens(EK100)~\cite{damen2022rescaling} and actions for Something-Something V2(SSV2)~\cite{goyal2017something}. For EGTEA~\cite{li2018eye}, in addition to top-1 accuracy, we include mean class accuracy, utilizing the initial train/test split, and for Charades-Ego~\cite{sigurdsson2018charades}, the evaluation metric is mean average precision (mAP).
The recognition accuracy for EK100 and SSV2 datasets are reported in Table~\ref{tab:EK100}(a) and Table~\ref{tab:SSV2}(b), respectively.
Our proposed method notably enhances the effectiveness of the existing supervised and self-supervised techniques for action recognition over ViT-B. We have at least $1.9\%$ improvement on action, verb, and noun on EK100 over the best results in the literature, AVION~\cite{zhao2023training} and LaViLa~\cite{zhao2023learning}. Similarly, on SSV2 action, the performance increase is $+0.9\%$ compared to the highest accuracy in the literature using ViT-B. To further evaluate its transferability, we use the FILS model that pretrained on EK100 and finetuned it on the Charades-Ego and EGTEA datasets, which are visually distinct from EK100. FILS improves mAP on Charades-Ego by $+1.3\%$ and top-1 accuracy on EGTEA by $+1.03\%$ compared to the baseline LAVILA model~\cite{zhao2023learning} after finetuning. Table~\ref{tab:chrades}(a) and Table~\ref{tab:EGTEA}(b) demonstrate our superior performance compared to the current state-of-the-art on these datasets. Therefore, instead of relying on large-scale datasets like the Kinetics~\cite{kay2017kinetics,carreira2019short} dataset, which includes URL links for up to 650,000 video clips, our method improved performance even when trained on a approximately 10x smaller dataset.

\subsection{Ablation Study}\label{subsec:ablation}

\noindent \textbf{Self-supervision Strategy}\label{subsec:str}
The core insight of FILS is to perform masked feature prediction in language semantic space. In Sec.~\ref{subsec:training-obj}, we introduce our initial objective, FP, which entails predicting features. 
To demonstrate the benefits of predicting patch features instead of reconstructing missing patches in the pixel domain using mean square error loss, we train a pair of ViT-B/16 models on the EK100 dataset using feature prediction loss and mean-squared error loss without the contrastive language contribution. To evaluate the action recognition accuracy of models trained with FP and MSE objectives, we finetuned a pretrained model on EK100. The model trained with the FP objective exhibits superior action accuracy, $50.3\%$, compared to the model trained with MSE, $48.5\%$. Both are lower than the $51.0\%$ achieved on EK100 action recognition via our full FILS models, evidence of the effectiveness of our proposed method, which involves feature prediction within the language space.

\begin{table}[t!]
      \caption{\textbf{Ablation study on two patch-wise contrastive scenarios and masking ratio.}}
      \vspace{0.2cm}
    \resizebox{0.49\linewidth}{!}
    {
        \begin{minipage}[h]{0.6\linewidth}
        \raggedleft
        \centering
                \phantom{hellohellohe}(a) ActCLIP strategies
            \begin{tabular}{cc cc}
                \toprule
                Method &  strategy & number of iteration & Action Top-1 Acc.\\
                \midrule
                FILS& patch &10& 38.0 \\
                FILS& patch &30& 43.5\\
                FILS& patch &50& 46.3\\
                FILS& patch-average &1& 51.0 \\
              \bottomrule
            \end{tabular}
        \end{minipage}  
     \label{tab:ActCLIP-ab}

    }
    \hspace{1.5cm}
    \begin{minipage}[h]{0.4\textwidth}
    \centering
     \phantom{hl}(b) Masking ratio
        \begin{tikzpicture}[scale=0.48]
            \begin{axis}[
                title={},
                xlabel={Masking ratio (\%)},
                ylabel={Top-1 Accuracy on action recognition (\%)},
                xmin=65, xmax=100,
                ymin=49, ymax=51.5,
                xtick={60,65,70,75,80,85,90,95,100},
                ytick={49,49.5,50,50.5,51,51.5},
                legend pos=north west,
                ymajorgrids=true,
                grid style=dashed,
            ]
            \addplot[
                color=blue,
                mark=square,
                nodes near coords,
                ]
                coordinates {
                (70,49.45)(75,49.8)(90,51.0)(95,50.4)
                };
                \legend{Epic-Kitchens}
            \end{axis}
        \end{tikzpicture}
    \end{minipage}

   \label{tab:masking-ab}
   \vspace{-0.6cm}
\end{table}

\begin{figure}[b!]
\vspace{-0.2cm}
  \centering
  \includegraphics[height=5.2cm]{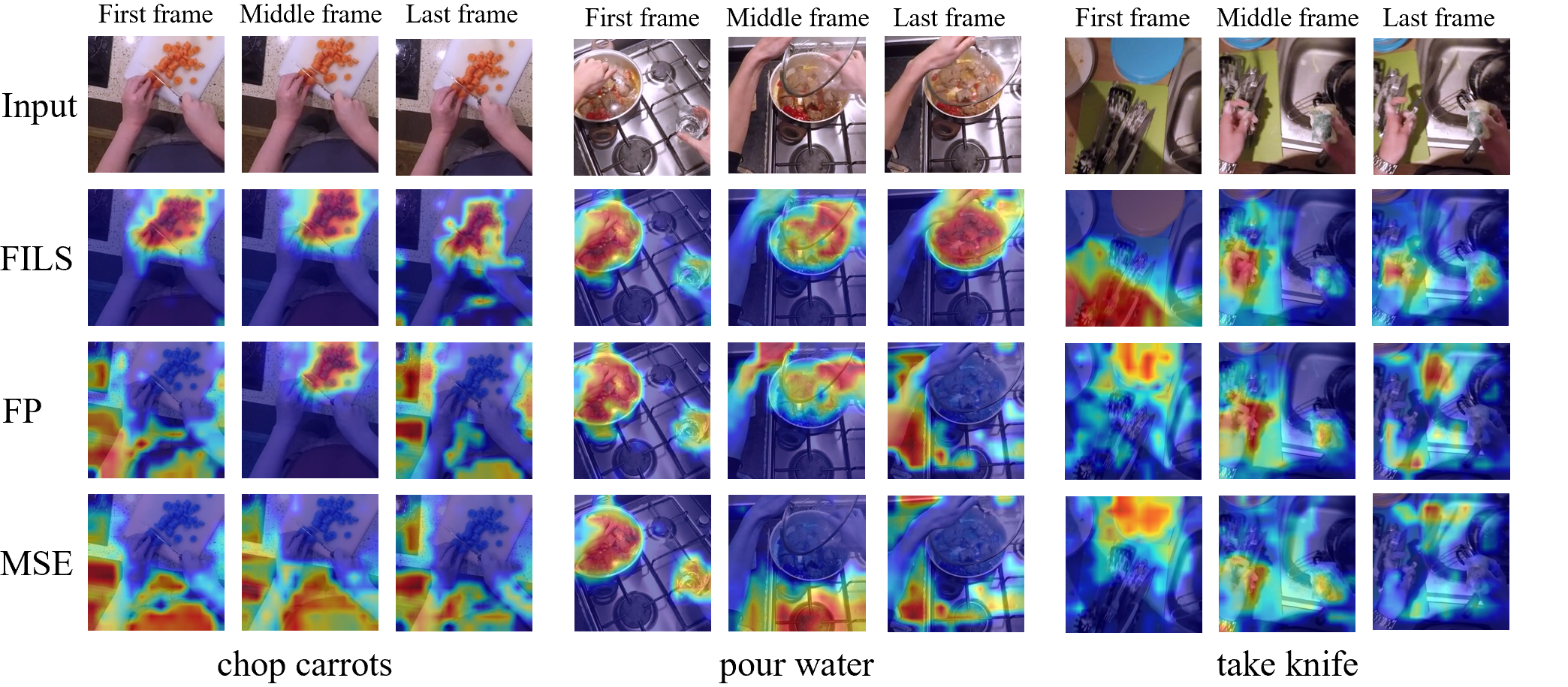}
  \caption{Attention heatmaps generated for the initial, central, and final frames of the EK100 using the last transformer layer of the model trained with self-supervised strategies including FILS, our second objective (FP), and pixel-domain reconstruction (MSE) after masking.
  }
  \label{fig:gradcam}
  \vspace{-0.4cm}
\end{figure}

\noindent\textbf{ActCLIP strategies.}
In our patch-wise ActCLIP framework, we applied contrastive learning between the average feature extracted from patches within the action area and text features. To examine the impact of the number of patches used, we conducted an ablation study (Table~\ref{tab:ActCLIP-ab}(a)). This involved randomly selecting one patch within the action area (patch strategy) instead of averaging across all patches within the action area (patch-average strategy). We repeated this process several times and documented the results alongside the number of iterations in Table~\ref{tab:ActCLIP-ab}(a). As anticipated, increasing the number of iterations improved performance, approaching the patch-average strategy utilized in ActCLIP.

\noindent\textbf{Masking Ratio.} We compare different masking ratios in Table~\ref{tab:masking-ab}(b) on our proposed method using Epic-Kitchens dataset. Increasing the masking ratio from $70\%$ to $95\%$ for tube masking, we find that the performance is higher with an extremely high ratio of $90\%$.

\begin{figure}[t!]
  \centering
  \includegraphics[trim={8cm 5cm 8cm 0cm}, height=5.2cm]{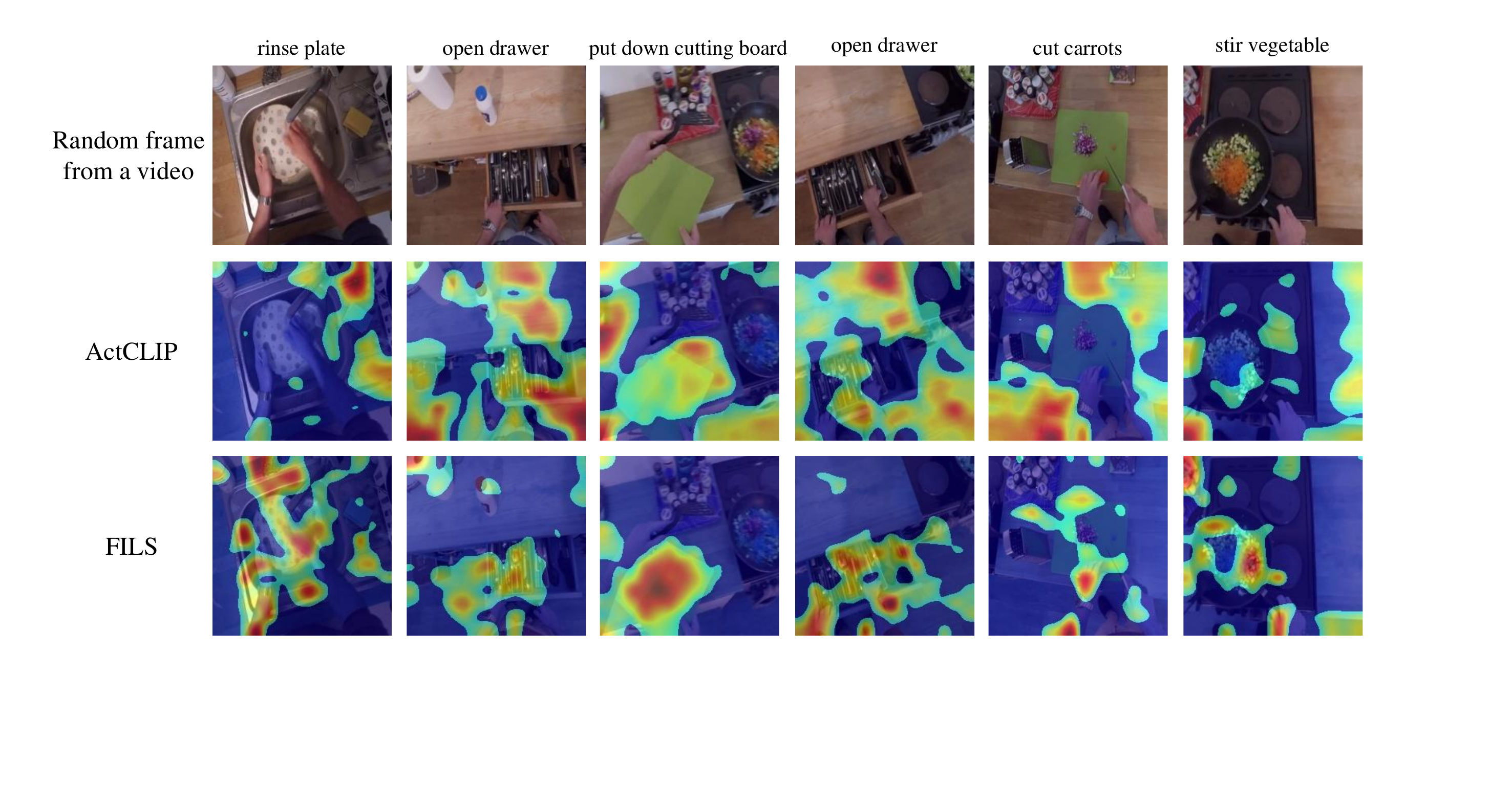}
  \caption{visualization of the similarity between text and video features for EK100 dataset. The provided text is the action label of the video we used.
  }
  \label{fig:FILSrep}
\end{figure}
\squeezeup

\subsection{Attention Visualization}

To gain deeper insight into the learned representations by FILS, we employ Grad-CAMs~\cite{selvaraju2017grad} to visualize the prominent areas that significantly contribute to the accomplishment of the action recognition task. This visualization helps us better comprehend the spatiotemporal cues acquired during the self-supervised learning step.
In Fig.~\ref{fig:gradcam}, we illustrate attention visualization for a few sample videos selected from the Epic-Kitchens-100 dataset; visualize attention heatmaps of the first, middle, and last frame of the video using the models trained with these training strategies: our proposed FILS, our first objective, which is FP, and MSE in the pixel domain. Sec.~\ref{subsec:str} discovers the comparison among these strategies. We selected instances that FILS correctly classified, whereas feature prediction (FP) and mean squared error (MSE) failed to do so. More attention visualizations on other datasets can be found in the supplemental material.
Our visualization study reveals that compared to the attention heatmaps computed by ViTs trained with MSE and FP objectives, ViTs trained with FILS produce attention heatmaps that emphasize the area where the action happens and are more effective at classifying the videos' actions. These findings further demonstrate the effectiveness of conducting contrastive learning between the patches within the action area and the relevant text in our proposed self-supervised technique (FILS).

\subsection{FILS learns semantic representations}
Our proposed approach predicts visual features in language space constructed through contrastive learning on patches within the identified action area alongside its corresponding text, thus receiving implicit guidance from the textual context. Using videos from the EK100, we calculate the similarity of features between the video patches and corresponding text features, which is the action label. Fig.~\ref{fig:FILSrep} compares our proposed contrastive strategy (first objective-ActCLIP) and our FILS. 
 FILS significantly enhances the contrastive objective between language and vision. This is evident in the heatmap's improved localization and reduced noise, which now exhibits greater concentration around the object and action regions.
To improve visualization, we smoothed the patch-wised attention blocks with Gaussian blurring.
 \squeezeup

%% file: tex/conclusion.tex
In this work, we introduce FILS, a novel self-supervised strategy combining two objectives: 1) ActCLIP - expanding image-based CLIP to video by capturing long-range temporal dependencies between frames using video-text contrastive loss in detected action areas, and 2) Feature prediction in the language semantic space built by ActCLIP. These complementary objectives yield synergistic benefits.
Experiments show FILS effectively solves downstream tasks like video action recognition, which require precise comprehension of motion. Due to constraints, we evaluated smaller datasets using ViT-B. Still, following prior research demonstrating contrastive learning's benefits from scale, we believe that using larger datasets containing millions of examples and a higher version of the vision transformer will further boost performance.

%% file: tex/supp.tex
\begin{center}
    \huge
    \textcolor{bmv@sectioncolor}{\textbf{Supplementary Material}}

\end{center}

In this supplementary material, we share implementation details and preprocessing step in Sec.~\ref{imp} and details of the datasets and evaluation metrics used in this work in Sec.~\ref{datasets}. We provide additional quantitative and qualitative experiments to enhance comprehension of the FILS in sections ~\ref{comp} to~\ref{cap}.

\section{Implementation Details}\label{imp}
For self-supervision, we sample 16 RGB frames from each video as a clip with a dynamic stride (depending on the number of raw video frames). The resolution of the frames is $224\times224$ and uses random resized cropping as augmentation. Our spatial patch size is 16×16 with a temporal patch size of 2. Therefore, each clip is split into non-overlapping 8×14×14 tubes, yielding 1568 tokens. The encoder in each of our experiments is the ViT-B/16 architecture~\cite{dosovitskiy2020image}, trained on Epic-Kitchens (EK100) and Something-Something V2 (SSV2) datasets. We use the AdamW optimizer~\cite{loshchilov2017decoupled} with $(\beta_1,\beta_2)=($0.9$, $0.95$)$ and a weight decay of 0.05. The learning rate starts at 1e-6, grows linearly to a peak of 1.5e-4 in the first epoch, and then uses a half-wave cosine schedule to decline to 1e-5 progressively. The predictor has six additional transformer layers. As a preprocessing step, we create synthetic captions for the whole videos in the training set using a video-to-text model (VideoBLIP)~\cite{Yu_VideoBLIP}. Our text encoder employs a CLIP-based encoder with frozen weights from ViFi-CLIP \cite{rasheed2023fine}. The pretraining is conducted for 800 epochs, and following~\cite{zhao2023training}, we employ flash attention~\cite{dao2022flashattention} to lessen the memory bottleneck of the attention operations; it is more efficient than standard attention techniques in terms of IO complexity. We also leverage gradient check pointing~\cite{chen2016training,child2019generating} for the training transformer to reduce the memory cost and use Pytorch gradient checkpointing.
For ViT-B, we employ a batch size of 64 per GPU over 4 GPUs, a total batch size of 256, significantly smaller than that used in previous studies in the order of thousands. 

We fix the parameters for all our finetuning experiments using the same hyperparameters as the trained baselines. We use AdamW with a momentum of 0.9 and weight decay of 0.05 to finetune the pretrained model on EK100 and SSV2 for a specific number of epochs on EK100, Charades-Ego, EGTEA, and SSV2. We employ cosine annealing with a warm-up, in which the base learning rate begins at 1e-6, increases linearly to a peak of 1.5e-3 in the first epoch, and then decreases gradually to 1e-6 using a half-wave cosine schedule. We replace the linear projection head for action classification with a dataset-specific dimension head. We marginalize the action-level probability to obtain verb- and noun-level accuracy. We also employ 0.8 mixup and 0.1 label smoothing. We input 16 sampled frames for each video clip during training and testing and resized the shorter side to 256 pixels. Next, we take a 224x224 crop and apply data augmentation using conventional Random Resized Crop (0.08,1.0) and Horizontal Flip (0.5), fused at the video-decoding side. We take the centre 224×224 crop at inference and scale the shorter side to 224 pixels. For ViT-B, we employ a batch size of 64 per GPU over 4 GPUs.

\section{Datasets and Metrics}\label{datasets}
To illustrate the performance of our approach, we apply the proposed method to four datasets: Something-Something V2(SSV2)~\cite{goyal2017something}, which is hard to discriminate the action classes for its videos because the same objects (and human hands) and very similar motions appear in many different actions, Epic-Kitchens that faces many challenges to predict its first-person activities due to limited field of view, occlusions and camera movements~\cite{damen2022rescaling}, Charades-Ego~\cite{sigurdsson2018charades} and EGTEA~\cite{li2018eye} which are relatively small datasets to validate the transferability of our pretrained model on them. 

\textbf{Epic-Kitchens}(EK100)~\cite{damen2022rescaling} is one of the largest egocentric (first-person) vision datasets, consisting of 700 variable-length videos over 100 hours, 20 million frames; each video is divided into short action segments, with a mean duration of 3.12 seconds, accompanied by annotations consisting of a verb and a noun describing the depicted action (for example, 'open cupboard'). With 67,217 in training, 9,668 in tests with 97 verbs, 300 nouns, and 3806 action classes.

\textbf{Something\text{-}Something V2}(SSV2)~\cite{goyal2017something}, this dataset depicts individuals in everyday human-object interaction tasks. The dataset is divided in an 8:1:1 ratio into train, validation, and test sets. With 168,913 videos in the training set, 24,777 in the validation set, and 27,157 in the test set, it has 220,847 videos and 174 labels.

\textbf{Charades-Ego}~\cite{sigurdsson2018charades}, the Charades-Ego dataset comprises 7,860 first- and third-person recordings of everyday indoor activities for 157 action classes. There are 34.4 hours of first-person and 34.4 hours of third-person recordings, with an average of 8.72 activities (68,536 activity instances) in each video. The videos are from 112 different rooms all over the world. We use the subset containing first-person data, consisting of 3,085 videos for training and 846 videos for testing purposes. 

\textbf{Extended GTEA Gaze+}(EGTEA)~\cite{li2018eye}, this dataset contains 29 hours of first-person videos from 86 sessions, from 32 participants completing seven meal preparation tasks in a realistic kitchen environment. The dataset includes action annotations for 10321 cases across 106 classes with an average duration of 3.2 seconds.

\section{Comparsion FILS with Pixel Reconstruction}\label{comp}

We demonstrate that our proposed method (FILS) exhibits notable enhancements, surpassing the performance of the pixel space reconstruction technique~\cite{tong2022videomae,zhao2023training}, which attained its results relying on extensive epochs of pertaining. We present the action recognition results of FILS and the pixel-reconstruction baseline on the Epic-Kitchens dataset using a ViT-B/16 pretrained on the self-supervised objective for 100, 400, and 800 epochs. This experiment aims to demonstrate our results' consistency and superior FILS performance over the pixel space reconstruction method, even with fewer training epochs.

\begin{figure}[htpb]
\vspace{0.5cm}
\centering
    \begin{tikzpicture}[scale=0.9]
        \centering
        \begin{axis}[
            title={},
            xlabel={epochs},
            ylabel={Top-1 Accuracy on action recognition (\%)},
            xmin=0, xmax=900,
            ymin=31, ymax=60,
            xtick={0,200,400,600,800,1000},
            ytick={31,33,35,37,39,41,43,45,47,49,51,53,55},
            legend pos=north west,
            ymajorgrids=true,
            grid style=dashed,
        ]
        
        \addplot[
            color=blue,
            mark=square,
            nodes near coords,
            ]
            coordinates {
            (100,36.2)(400,41.0)(800,51.0)
            };
    
            \addplot[
            color=red,
            mark=triangle,
            nodes near coords,
            every node near coord/.append style={xshift=10pt,yshift=-8pt,anchor=east,font=\footnotesize},
            ]
            coordinates {
            (100,33.7)(400,40.5)(800,48.5)
            };
        \legend{feature prediction in language space, pixel space reconstruction}
        \end{axis}
        
    \end{tikzpicture}
\caption{The impact of varying pretraining epochs on the Epic-Kitchens-100 dataset. There is a consistent upward trend in action recognition accuracy with an increase in the number of pretraining epochs.}
\end{figure}
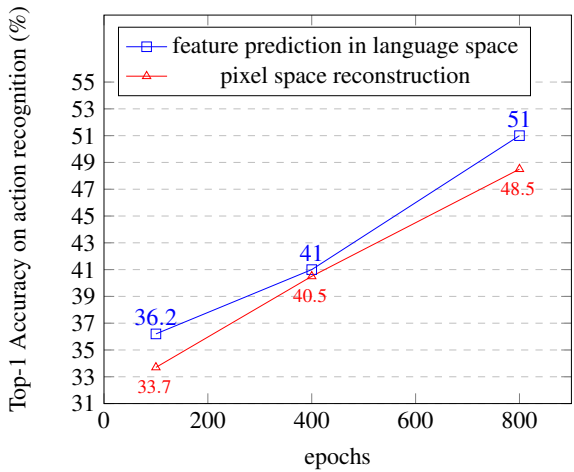

\section{Charades-Ego and EGTEA Action Recognition using FILS is pretrained on SSV2}

To further assess the transferability of our proposed FILS, we trained our FILS on Epic-Kitchens and then finetuned it on Charades-Ego and EGTEA datasets; reported results in Sec.4.1 in the main submission demonstrate our superior performance compared to the current state-of-the-art on these datasets. Additionally, we have extended our evaluation by incorporating action recognition results for Charades-Ego and EGTEA using FILS pretrained on the SSV2 dataset. As shown in Tables~\ref{tab:chrades} and~\ref{tab:EGTEA}, the FILS model pretrained on SSV2 also achieved superior performance, outperforming the previous state-of-the-art results. These findings underscore the transferability of FILS across different datasets, highlighting its potential for broader application in action recognition tasks.

\begin{table}[htb]
\centering

  \caption{\textbf{Charades-Ego Action Recognition.} FILS achieves substantial improvements in this task, surpassing prior works in both scenarios:  FILS undergoes pretraining on either EK100 or SSV2.
In the table, 'p-data' signifies the pretraining data employed, while 'L' denotes the incorporation of language through the training process.}
\vspace{5mm}

\label{tab:othersFT}
    \resizebox{9.5cm}{!}{
    \raggedleft
  \centering
  \scalebox{0.65}{
  \begin{tabular}{c c c c c}
    \toprule
    Method & Backbone & p-data & L & mAP\\
    \midrule
    ActorObserverNet~\cite{sigurdsson2018actor} & ResNet-152 & Charades & $\times $ &20\\
    SSDA~\cite{choi2020unsupervised} & I3D & Charades-Ego& $\times$ &25.8\\
    Ego-Exo~\cite{li2021ego}  & SlowFast-R101  & Kinetics-400 & $\times$ &30.1\\
    EgoVLP~\cite{lin2022egocentric} & TSF-B & Ego4D & $\checkmark$ &32.1 \\
    HierVL-Avg~\cite{ashutosh2023hiervl} &ViT-Base& Ego4D&$ \checkmark $ & 32.6 \\
    HierVL-SA~\cite{ashutosh2023hiervl} & ViT-Base & Ego4D & $ \checkmark $& 33.8\\
    EgoVLPv2~\cite{pramanick2023egovlpv2} & TSF-B & EgoClip & $\checkmark$ & 34.1\\
    LaViLA~\cite{zhao2023learning} & TSF-B &  WIT+Ego4D & $ \checkmark $ &33.7\\

    \textbf{FILS(ours)} & ViT-Base & EK100 & $ \checkmark $ & \textbf{34.4} \\
    \textbf{FILS(ours)} & ViT-Base & SSV2 & $ \checkmark $ & \textbf{34.2} \\
  \bottomrule
  \end{tabular}
  }
   \label{tab:chrades}
 
    }
\vspace{1cm}

\end{table}

\begin{table}[htb]
  \caption{\textbf{EGTEA Action Recognition.} FILS significantly enhances performance in this task, outperforming previous methods in both cases where it is pretrained on either EK100 or SSV2. In the table, 'p-data' represents the pretraining dataset used, and 'L' indicates the inclusion of language during the training process.}
\vspace{5mm}

\label{tab:othersFT}
    \centering
    \resizebox{10cm}{!}{
  \centering
  
    \scalebox{0.65}{
  \begin{tabular}{c c c c c c}
    \toprule
    Method & Backbone & p-data & L & Top-1 Acc. & Mean Acc.\\
    \midrule
    Li et al.~\cite{li2018eye} & I3D  & K400 & $\times$ & - & 53.30\\
    LSTA~\cite{sudhakaran2019lsta}  & ConvLSTM & IN-1k &  $\times$ & 61.86 & 53.00\\
    IPL~\cite{wang2021interactive} & I3D & K400 & $\checkmark$ & - & 60.15\\
    MTCN~\cite{kazakos2021little} & SlowFast  & K400+VGG-S  & $ \checkmark $& 73.59 & 65.87\\
    LaViLA~\cite{zhao2023learning}  & TSF-B & WIT+Ego4D & $ \checkmark $ & 77.45 & 70.12\\
    \textbf{FILS(ours)} & ViT-Base & EK100 & $ \checkmark $   & \textbf{78.48} & \textbf{71.20} \\
    \textbf{FILS(ours)} & ViT-Base & SSV2 & $ \checkmark $   & \textbf{78.57} & \textbf{71.31} \\

  \bottomrule
  \end{tabular}
  }
      \label{tab:EGTEA}
    }

\end{table}
\section{Qualitative Results}
\begin{figure}[t!]

  \centering
  \includegraphics[trim={1cm -15cm 10cm -7cm}, height=22cm]{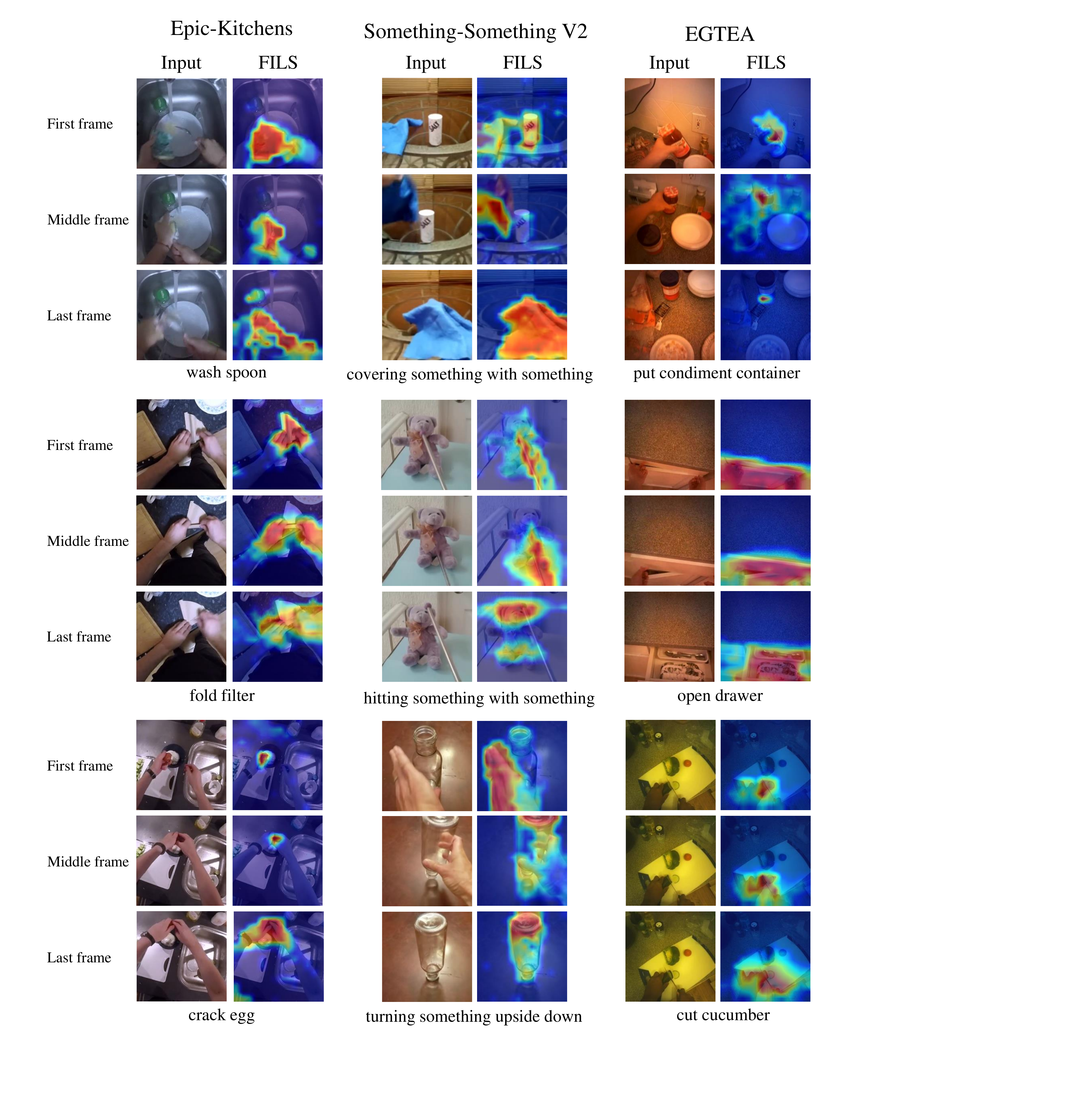}
  \vspace{-4.5cm}

    \caption{Visualization by Grad-CAM on Epic-Kitchens, Something-Something V2 and EGTEA.
  }
  \label{fig:FILSrep}
\end{figure}

We use Grad-CAM~\cite{selvaraju2017grad} to visualize the last stage feature maps.
In Fig.~\ref{fig:FILSrep}, we display the Grad-CAM of the first, middle, and last frames for the challenging examples from the datasets we worked on: Epic-Kitchens, Something-Something V2, and EGTEA. The attention maps demonstrate how well our proposed self-supervised technique (FILS) represents the potential semantic region in the video and acquires an understanding of spatiotemporal relationships by linking pertinent areas, revealing distinctly the semantics of actions. 

\begin{figure}[htb!]
  \centering
  \includegraphics[trim={8cm 3cm 8cm 1cm}, height=5.5cm]{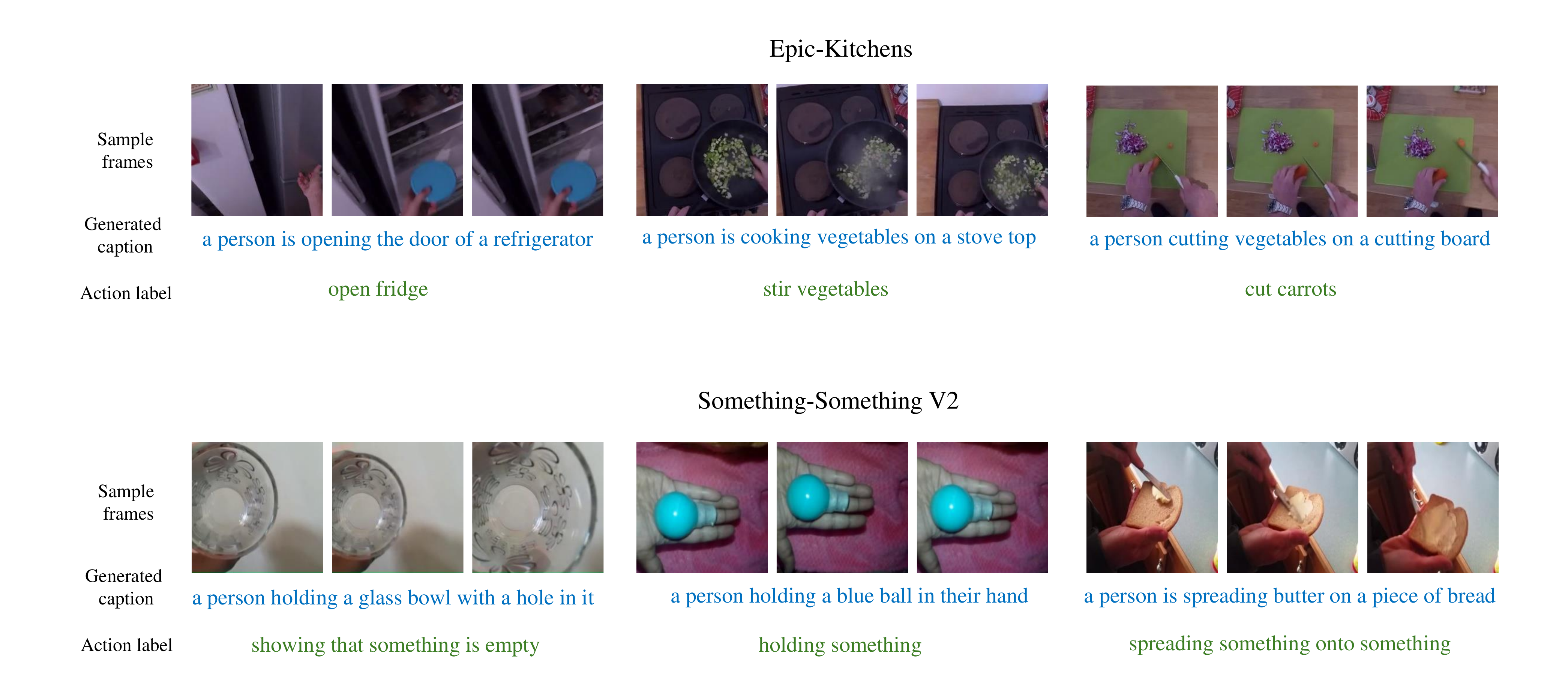}
  \vspace{5mm}

  \caption{ Synthetic captions for some instances from the training set of Epic-Kitchens and Something-Something v2. VideoBLIP often captures good spatial and temporal details. 
  }
  \label{fig:caption}
\end{figure}

\section{Synthetic Captions}\label{cap}
In our self-supervision step, as a preliminary step, we use a video-to-text model (VideoBLIP)~\cite{Yu_VideoBLIP} to generate synthetic captions for all of the videos in the training set. This process was conducted for both the Epic-Kitchens and Something-Something v2 datasets. In Fig.~\ref{fig:caption}, we present examples of generated captions alongside their respective labels for the training sets of these two datasets. These synthetic captions for videos have demonstrated remarkable effectiveness, providing detailed scene descriptions and capturing concepts closely related to video labels, which could enhance the interpretability of video content. Using these generated captions in our training process could enhance the comprehension and performance of our model, as these enriched input features are semantically meaningful, which is crucial for training models.

